\pdfoutput=1
\documentclass[11pt]{article}

\usepackage[]{acl}

\usepackage{times}
\usepackage{latexsym}

\usepackage[T1]{fontenc}
\usepackage[utf8]{inputenc}

\usepackage{microtype}

\usepackage{booktabs}
\usepackage{amsmath}
\usepackage{graphicx}
\usepackage{enumitem}

\usepackage[skip=0pt,belowskip=0pt]{caption}
\everypar{\looseness=-1}
\title{On Systematic Style Differences between Unsupervised and Supervised MT \\
and an Application for High-Resource Machine Translation}
\author{Kelly Marchisio\thanks{\hspace{2mm}Work completed at Google Translate Research.} \\ Johns Hopkins University \\ kmarc@jhu.edu \\
    \And Markus Freitag \hspace{4mm}  David Grangier \\ Google Research \\ \{freitag, grangier\}@google.com}

\begin{document}
\maketitle
\begin{abstract}
Modern unsupervised machine translation systems reach reasonable translation quality under clean/controlled data conditions. As the performance gap between supervised and unsupervised MT narrows, it is interesting to ask whether the different training methods result in systematically different output beyond what is visible via quality metrics like adequacy or BLEU. We compare translations from supervised and unsupervised MT systems of similar quality, finding that unsupervised output is more fluent and more structurally different in comparison to human translation than is supervised MT. We then demonstrate a way to combine the benefits of both methods into a single system which results in improved adequacy and fluency as rated by human evaluators. Our results open the door to interesting discussions about how supervised and unsupervised MT might be different yet mutually-beneficial. 
\end{abstract}

\section{Introduction}
Supervised machine translation (MT) utilizes parallel bitext to learn to translate. Ideally, this data consists of natural texts and their human translations.  In a way, the goal of supervised MT training is to produce a machine that mimicks human translators in their craft. 
Unsupervised MT, on the other hand, uses monolingual data alone to learn to translate.  Critically, unsupervised MT \textit{never sees an example of human translation}, and therefore \textit{must create its own style of translation}. Unlike supervised MT where one side of each training sentence pair must be a translation, unsupervised MT can be trained with natural text alone.

In this work, we investigate the output of supervised and unsupervised MT systems of similar quality to assess whether systematic differences in translation exist.  Our exploration of this research area focuses on English$\to$German for which abundant bilingual training examples exist, allowing us to train high-quality systems with both supervised and unsupervised training.

\vspace{4pt}
Our main contributions are:
% \begin{itemize}[noitemsep,topsep=0pt]
\begin{itemize}[topsep=3pt]
\item We observe systematic differences between the output of supervised and unsupervised MT systems of similar quality. High-quality unsupervised output appears \textbf{more natural}, and more \textbf{structurally diverse} when compared to human translation.
\vspace{-0.5em}
\item We show a way to incorporate unsupervised back-translation into a standard supervised MT system, improving adequacy, naturalness, and fluency as measured by human evaluation.
\end{itemize}

Our results provoke interesting questions about what unsupervised methods might contribute beyond the traditional context of low-resource languages which lack bilingual training data, and suggest that unsupervised MT might have contributions to make for high-resource scenarios as well. It is worth exploring how combining supervised and unsupervised setups might contribute to a system better than either creates alone.

We discuss related work in \S\ref{sec:related_work}. In \S\ref{sec:exp_seup}, we introduce the dataset, model details, and evaluation setups. In \S\ref{sec:differences}, we characterize the differences between the output of unsupervised and supervised neural MT systems of similar quality. In \S\ref{sec:joined_mt}, we demonstrate a combined system which benefits from the complementary strengths of the two methods. We summarize the paper in \S\ref{sec:conclusion}.

\section{Related Work}
\label{sec:related_work}
\paragraph{Unsupervised MT}
Two paradigms for unsupervised MT are finding a linear transformation to align two monolingual embedding spaces~\cite{DBLP:conf/iclr/LampleCDR18, lample-phrase-2018, conneau-lample-2018, artetxe2018iclr, artetxe2019}, and pretraining a bi-/multilingual language model then finetuning on a translation task~\cite{conneaulample2019cross, song2019mass, liu2020multilingual}. We study the Masked Sequence-to-Sequence Pretraining (MASS) language model pretraining paradigm of~\citet{song2019mass}.  MASS is an encoder-decoder trained jointly with a masked language modeling objective on monolingual data. Iterative back-translation (BT) follows pretraining. 

\paragraph{Monolingual Data in MT}
BT is widely-used to exploit monolingual data \cite{sennrich2016improving}. ``Semi-supervised'' systems use monolingual and parallel data to improve performance (e.g.~\citet{artetxe2018iclr}). \citet{siddhant2020leveraging} combine multilingual supervised training with MASS for many languages and zero-shot translation.

\paragraph{Source Artifacts in Translated Text}
Because supervised MT is trained ideally on human-generated translation, characteristics of human translation affects the style of machine translation from such systems. Dubbed ``translationese,'' human translation includes source language artifacts \cite{koppel2011translationese} and source-independent artifacts---{\it Translation Universals} \cite{mauranen2004translation}. There are systematic biases inherent to translated texts~\cite{baker1993corpus, selinker1972interlanguage}, and biases coming from interference from source text~\cite{toury1995descriptive}. In MT,~\citet{freitag-etal-2019-ape,freitag-etal-2020-bleu} attribute these patterns as a source of mismatch between BLEU~\cite{papineni2002bleu} and human evaluation measures of quality, raising concerns that overlap-based metrics reward hypotheses with the characteristics of translated text more than those with natural language. \citet{vanmassenhove-etal-2019-lost, vanmassenhove2021machine} note loss of linguistic diversity and richness from MT. \citet{toral-2019-post} see related effects even after human post-editing.
The impact of translated text on human evaluation has been studied \citep{Toral18,Zhang19,Graham19,fomicheva-specia-2016-reference,ma-etal-2017-investigation}, as has the impact in training data \cite{kurokawa09,Lembersky12adapting,bogoychev2019domain,riley-etal-2020-translationese}.

\paragraph{Measuring Word Reordering}
Word reordering models are well-studied because they formed a critical part of statistical MT (see \citet{bisazza2016survey} for a review). Others examined metrics for measuring reordering in translation \cite[e.g.][]{birch-etal-2008-predicting, birch2009quantitative, birch2010metrics}. \citet{wellington-etal-2006-empirical} and \citet{fox-2002-phrasal} use part-of-speech (POS) tags in the context of parse trees, and \citet{fox-2002-phrasal} measure the similarity of French and English with respect to phrasal cohesion by calculating alignment crossings using parse trees. Most similar to us, \citet{birch2011reordering} view simplified word alignments as permutations and compare distance metrics over these to quantify the amount of reordering done. They use TER computed over the alignments as a baseline. \citet{birch2011reordering-acl}'s LRScore interpolates a reordering metric with a lexical translation metric. 

\section{Experimental Setup}
\label{sec:exp_seup}

\subsection{Data}
\paragraph{Training}
Experiments are in English$\to$German. For the main study comparing supervised and unsupervised MT, we use News Commentary v14 (329,000 sentences) as parallel bitext for the supervised system, and News Crawl 2007-17 as monolingual data for the unsupervised system. Deduplicated News Crawl 2007-17 has 165 million English sentences and 226 million German sentences.
% News Commentary is human-translated data crawled from https://www.project-syndicate.org/
% Sources: personal communication w/ Philipp Koehn, email: 14 Jan 2022.
% Also: http://www.meta-net.eu/meta-research/publications/publications/lrecannotatedcorpus.pdf
% Also: https://www.google.com/books/edition/Artificial_Intelligence_and_Natural_Lang/bSAAEAAAQBAJ?hl=en&gbpv=1&dq=casmacat+project-syndicate&pg=PA197&printsec=frontcover

The combined system demonstration at the end of our work utilizes a BT selection method. We use the bilingual training data from WMT2018~\cite{bojar-EtAl:2018:WMT1} (News Commentary v13, Europarl v7, Common Crawl, EU Press Release) so that our model can be compared with well-known work using BT \cite[e.g.][]{edunov2018understanding,caswell-etal-2019-tagged}.  We deduplicate and filter out pairs with $>250$ tokens in either language or length ratio over 1.5, resulting in 5.2 million paired sentences.

\paragraph{Development and Test Sets}
For the main experiments, we use newstest2017 as the dev set with newstest2018 and newstest2019 for test. newstest2018 was originally created by translating one half of the test data from English$\to$German (orig-en) and the other half from German$\to$English (orig-de). Since 2019, WMT produces newstest sets with only source-original text and human translations on the target side to mitigate known issues when translating and evaluating on target-original data \cite[e.g.][]{koppel2011translationese, freitag-etal-2019-ape}. 

For most experiments, we evaluate on orig-en sentences only to reflect the real use-case for translation and modern evaluation practice. We examine orig-de only for BLEU score as an additional data point of difference between supervised and unsupervised MT. \citet{Zhang19} show that target-language-original text should not be used for human evaluation (orig-de, in our case).  

We use the newstest2018 ``paraphrased'' test references from~\citet{freitag-etal-2020-bleu},\footnote{github.com/google/wmt19-paraphrased-references} which are made for orig-en sentences only. These additional references have different structure than the source sentence but maintain semantics, and provide a way to measure system quality without favoring translations with the same structure as the source. Observing work that uses these references, BLEU is typically much lower than on original test sets, and score differences tend to be small but reflect tangible quality difference \cite{freitag-etal-2020-bleu}.

For the system combination demonstration, we use newstest2018 for development and newstest2019 for test. We also use newstest2019 German$\to$English and swap source and target to make an orig-de English$\to$German test set, and use paraphrase references for newtest2019 (orig-en). 

Testing on the official newstest2018 in the main experiments allows us to see interesting differences between unsupervised and supervised MT that are hidden with newstest2019 because it is orig-en only. Using newstest2018 for development in the system combination demonstration aligns with similar literature \cite[e.g.][]{edunov2018understanding,caswell-etal-2019-tagged}. We use SacreBLEU throughout \cite{post-2018-call}.\footnote{BLEU+case.mixed+lang.ende+numrefs.1+smooth.exp+
\{TESTSET\}+tok.13a+version.1.4.12}

\subsection{Part-of-Speech Tagging}
We use part-of-speech taggers for some experiments:
universal dependencies (UD) implemented in spaCy\footnote{https://spacy.io/, https://universaldependencies.org/} and spaCy's language-specific fine-grained POS tags for German from the TIGER Corpus~\cite{albert2003tiger,brants2004tiger}.

\subsection{Models}
Our \textbf{unsupervised MT} model is a MASS transformer with the hyperparameters of~\citet{song2019mass}.  We train MASS on the News Crawl corpora, hereafter called ``Unsup.'' Our \textbf{supervised MT} systems use the transformer-big~\cite{vaswani_2017} as implemented in {\it Lingvo}~\cite{shen2019lingvo} with a vocabulary of 32k subword units.

To investigate differences between approaches, we train two \textbf{language models} (LMs) on different types of data and calculate the perplexity of translations generated by the supervised and unsupervised MT systems.
We train one LM on the monolingual German News Crawl dataset with a decoder-only transformer, hereafter called the ``natural text LM'' (nLM). We train another on machine translated sentences which we call the ``translated text LM'' (tLM). We generate the training corpus by translating the English News Crawl dataset into German with a English$\rightarrow$German transformer-big model trained on the WMT18 bitext.

\subsection{Human Evaluations}
\label{sec:huma_eval}
Human evaluation complements automatic evaluation and abstracts away from comparison to a human reference which favors the characteristics of translated text~\cite{freitag-etal-2020-bleu}.
We score adequacy using direct assessment and run side-by-side evaluations measuring fluency and adequacy preference between systems. Each campaign has 1,000 test items. For side-by-side eval, a test item includes a pair of translations of the same source sentence: one from the supervised system and one from the unsupervised. 
We hire 12 professional translators, who are more reliable than crowd workers~\cite{toral2020reassessing,freitag2021experts}.

\paragraph{Direct Assessment Adequacy}
We use the template from the WMT 2019 evaluation campaign. Human translators assess a translation by how adequately it expresses the meaning of the source sentence on a 0-100 scale. Unlike WMT, we report the average rating and do not normalize the scores. 
\paragraph{Side-by-side Adequacy}
Raters see a source sentence with two translations (one supervised, one unsupervised) and rate each on a 6-point scale. 
\paragraph{Side-by-side Fluency}
Raters assess the alternative translations (one supervised, one unsupervised) without the source, and rate each on a 6-point scale.

\section{Unsupervised vs. Supervised MT}
\label{sec:differences}
The goal of this section is to analyse supervised and unsupervised systems of similar overall translation quality so that differences in quality do not confound analyses. As unsupervised systems underperform supervised systems, we use a smaller parallel corpus (news commentary) to train systems of similar quality. 
Table~\ref{tab:system_perf} summarizes the BLEU scores and human side-by-side adequacy results for both systems.
Although the supervised system is below state-of-the-art, these experiments help elucidate how unsupervised and supervised output is different. 
Overall BLEU and human ratings suggest similar translation quality. Nevertheless, we observe notable differences between orig-de and orig-en sides of the test set when comparing both systems. Recall that orig-de has natural German text on the target side. Unsup scores higher than Sup on orig-de, suggesting that its output is more natural-sounding as it better matches text originally written in German. Performance discrepancies on orig-en and orig-de indicate that differences in system output may exist and prompt further investigation.

\begin{table}[h]
\vspace{-0.5em}
\small
\begin{center}
\resizebox{\columnwidth}{!}{%
\setlength{\tabcolsep}{2pt}
\begin{tabular}{l|ccc|c|c}
&\bf Overall & \bf orig-en & \bf orig-de & \bf nt18p & \bf Human Adq. \\ 
\hline
\bf Sup & 29.2 & 34.0 & 21.1 & 9.3 & 3.89 \\
\bf Unsup & 30.1 & 30.9 & 27.1 & 9.6 & 3.82 \\
\end{tabular}}%
\end{center}
\vspace{-0.5em}
\caption{\label{tab:system_perf} SacreBLEU \& human adequacy (orig-en) on newstest2018 and newstest2018p (nt18p = paraphrase reference).  nt18p is available for orig-en only.}
\vspace{-1.7em}
\end{table}

\subsection{Selecting Translations of Same Adequacy}
To assess the translation style and compare linguistic aspects of supervised and unsupervised MT, we further must compare translations that have the same accuracy level on the segment level, so that neither confounds analysis. We use the adequacy evaluation from Table~\ref{tab:system_perf} and retain sentences for which both approaches yield similar adequacy scores. We divide the rating scale into bins of low (0--2), medium (3--4), and high (5--6) adequacy. Table \ref{tab:human_eval_bins} shows the percentage of sentences in each bin. For each source sentence, there is one translation by Unsup and one by Sup. If human judges assert that both translations belong in the same adequacy bin, that sentence also appears in ``Both.'' There are 86, 255, and 218 sentences in ``Both'' for low, medium, and high bins, respectively. For subsequent analyses, we examine sentences falling into ``Both.'' 

\begin{table}[h]
\vspace{-0.5em}
\small
\begin{center}
\resizebox{\columnwidth}{!}{%
\setlength{\tabcolsep}{11pt}
\begin{tabular}{c|c|c|c}
& \bf Low & \bf Medium & \bf High \\ 
\hline
\bf Sup & 18.7\% & 42.1\% & 39.2\% \\
\bf Unsup & 19.3\% & 44.6\% & 36.1\% \\
\bf Both & 8.6\% & 25.5\% & 21.8\% \\
\end{tabular}}%
\end{center}
\vspace{-0.5em}
\caption{\label{tab:human_eval_bins} Percentage of sentences with low, medium, high human-evaluated adequacy ratings. ``Both'' are sentences which have same rating from both systems.}
\vspace{-1.7em}
\end{table}

\subsection{Comparing Translation Style}
\paragraph{Measuring Structural Similarity}
\label{sec:mono}
We develop a metric to ascertain the degree of structural similarity between two sentences, regardless of language. When evaluated on a source-translation pair, it measures the influence of the source structure on the structure of the output without penalizing for differing word choice;  thus it is a measure of ``monotonicity'' -- the degree to which words are translated in-order. Given alternative translations in the same language, it assesses the degree of structural similarity between the two. Thus given a machine translation and a human translation of the same source sentence, it can measure the structural similarity between the machine and human translations. 

Word alignment seems well-suited here.  Like \citet{birch2011reordering}, we calculate Kendall's tau~\cite{kendall1938new} over alignments of source-translation pairs, but do not simplify alignments to permutations. We use fast\_align~\cite{dyer-etal-2013-simple} but observe that it struggles to align words not on the diagonal, so sometimes skipped alignments.\footnote{We ran fast\_align with and without diagonal-favoring and all 5 symmetrization heuristics, and see similar trends.} 
% This may make the correlation coefficient deceptively high. 
Because of this issue, we instead estimate monotonicity/structural similarity using the new metric, introduced next.

We propose measuring translation edit rate (TER, ~\citet{snover2006ter}) over \textit{POS tag sequences}.  TER is a well-known word-level translation quality metric which measures the number of edits required to transform a ``hypothesis'' sentence into the reference, outputting a ``rate'' by normalizing by sentence length.  Between languages, we compute TER between POS tag sequences of the source text (considered the reference) and the translation (considered the hypothesis), so that TER now measures changes in \textit{structure} independent of word choice. Source-target POS sequences which can be mapped onto each other with few edits are considered similar---a sign of a monotonic translation.  Given a machine translation (hypothesis) and a human reference in the same language, TER over POS tags measures \textit{structural similarity between the machine and human translations}. Outputs with identical POS patterns score 0, increasing to 1+ as sequences diverge. Lower TER for (source, translation) pairs indicates monotonic translation; Lower TER for (machine translation, human translation) pairs indicates structural similarity to human translation.  We call the metric ``posTER''.

\paragraph{Monotonicity}
POS sequences are comparable across languages thanks to universal POS tags. Table \ref{tab:pos} has a toy example with two possible German translations of an English source. Next to each sentence is its universal dependencies POS sequence.  In the third column, TER is calculated with the POS sequence of the English source as reference and the sequence of the translation as hypothesis.

\begin{table*}[h]
\small
\begin{center}
\resizebox{\textwidth}{!}{%
\setlength{\tabcolsep}{11pt} %default is 6pt
\begin{tabular}{l|l|l}
\textbf{Sentence} & \textbf{POS Sequence} & \textbf{TER} \\ \hline
\textit{I made myself a cup of coffee this morning.}& \textit{PRON VERB PRON DET NOUN ADP} & \textit{-} \\ 
& \textit{PNOUN DET NOUN PUNCT} &  \\
\hline
Ich habe mir heute Morgen eine Tasse & PRON AUX PRON ADV NOUN DET & 0.5 \\ 
Kaffee gemacht. & NOUN NOUN VERB PUNCT & \\
\hline
Heute morgen habe ich mir eine Tasse &ADV ADV AUX PRON PRON DET & 0.7 \\
Kaffee gemacht. & NOUN NOUN VERB PUNCT & 
\end{tabular}}%
\end{center}
\vspace{-0.5em}
\caption{\label{tab:pos}posTER over universal dependencies POS sequences for example toy German translations of an English source. Row 1 is the source with its POS tag sequence. Rows 2/3 are example translations with POS tags. posTER is calculated via the POS sequences of the translation (hypothesis) and the source (considered the reference).}
\vspace{-1.25em}
\end{table*}

\begin{table}[h]
\vspace{-0.25em}
\small
\begin{center}
\resizebox{\columnwidth}{!}{%
\setlength{\tabcolsep}{12pt}
\begin{tabular}{@{}c||c|c|c|c@{}}
\toprule
& \bf nt18 & \bf nt18p & \bf Sup & \bf Unsup \\
\hline
\bf Src & 0.410 & 0.546 & 0.409 & 0.399
\end{tabular}}%
\end{center}
\vspace{-0.5em}
\caption{\label{tab:ter_cross} posTER (0-1+) over universal dependencies of translations of newstest2018 (orig-en) vs. the source. $\downarrow$ = more monotonic translation. nt18p=paraphrase ref.}
\vspace{-1.5em}
\end{table}

Table \ref{tab:ter_cross} shows posTER over universal dependencies of German translations versus the newstest2018 (orig-en) source sentences. While the standard newstest2018 references (Ref) score 0.410, newstest2018p's (RefP) higher score of 0.546 reflects the fact that the paraphrase references are designed to have different structure than the source. Difference in overall monotonicity between Sup and Unsup is unapparent at this granularity. 

Because universal dependencies are designed to suit many languages, the 17 UD categories may be too broad to adequately distinguish moderate structural difference. Whereas UD has a single class for ``VERB,'' the finer-grained German TIGER tags distinguish between 8 sub-verb types including infinitive, modal, and imperative. We use these language-specific categories next to uncover differences between systems that broad categories conceal.

\paragraph{Similarity to Human Translation}
Recall that supervised MT essentially mimics human translators, while unsupervised MT learns to translate without examples. Intuitively, supervised MT output might be stylistically more like human translation, even when controlling for quality. The first indication is Sup's lower BLEU score on nt18p---the paraphrase test set designed to have structure different than the original human translation.

We compare the structure of MT output with the human reference using German TIGER tags. Lower posTER indicates more structural similarity, while higher posTER indicates stylistic deviation from human translation. Comparison with the newstest2018 orig-en human reference is in Table \ref{tab:ter_qual}. Sup and Unsup show negligible difference overall, but binning by adequacy shows Unsup output as \textit{less structurally similar to the human reference on the high-end of adequacy}, and more similar on the low-end. This suggests systematic difference between systems, and that unsupervised MT might have more structural diversity as quality improves. 

\begin{table}[h]
\vspace{-0.25em}
\small
\begin{center}
\resizebox{\columnwidth}{!}{%
\setlength{\tabcolsep}{8pt}
\begin{tabular}{c||c|ccc}
\toprule
& \bf Overall & \bf Low & \bf Med & \bf High\\ 
\hline
\bf Sup & 0.280 & 0.348 & 0.282 & 0.255 \\
\bf Unsup & 0.287 & 0.313 & 0.298 & 0.296
\end{tabular}}%
\end{center}
\vspace{-0.5em}
\caption{\label{tab:ter_qual} posTER (0-1+ scale) over TIGER POS tags of system output vs. the human reference, grouped by adequacy (newstest2018, orig-en). $\downarrow$ = greater structural similarity to the human reference.}
\vspace{-1.25em}
\end{table}

\paragraph{Naturalness}
The first hint that Unsup might produce more natural output than Sup is its markedly higher BLEU on the orig-de test set: 27.1, versus 21.1 from Sup. Recall that orig-de has natural German on the target side, so higher BLEU here means higher n-gram overlap with natural German.

\citet{edunov2020evaluation} recommend augmenting BLEU-based evaluation with perplexity from a language model (LM) to assess fluency or naturalness of MT output. Perplexity \cite{jelinek1977perplexity} measures similarity of a text sample to a model's training data. We contrast the likelihood of output according to two LMs: one trained on machine-translated text (tLM) and another trained on non-translated natural text (nLM). While machine-translated and human-translated text differ, the LMs are nonetheless a valuable heuristic and contribute insights on whether systematic differences between MT system outputs exist. Low perplexity from the nLM indicates natural language. Low perplexity from the tLM (trained on English News Crawl that has been machine-translated into German) shows proximity to training data composed of translated text, indicating simplified language. 

Sup perplexity is lower than Unsup across adequacy bins for the tLM, seen in Table \ref{tab:lm}. Conversely, Sup generally has higher perplexity from the nLM. All adequacy levels for Unsup have similar nLM perplexity, suggesting it is particularly skilled at generating fluent output. Together, these findings suggest that \textit{unsupervised MT output is more natural} than supervised MT output. 

\begin{table*}[h]
\vspace{-0.5em}
\small
\begin{center}
\resizebox{\textwidth}{!}{%
\setlength{\tabcolsep}{10pt}
\begin{tabular}{c||c|ccc||c|ccc}
&  \multicolumn{4}{c||}{\bf Natural Text LM \hspace{0.5em} $\downarrow$}  & \multicolumn{4}{c}{\bf Translated Text LM \hspace{0.5em} $\uparrow$}  \\
\hline
& \it Overall & \it Low & \it Medium & \it High   & \it Overall & \it Low & \it Medium & \it High \\ 
\hline
\bf Sup & 72.69 & 90.61 & 76.36 & \textbf{68.37} & 41.06 & 51.91 & 40.32 & 36.70   \\
\bf Unsup & \textbf{67.01} & \textbf{68.32} & \textbf{60.56} & 69.88  & \textbf{58.17} & \textbf{61.50} & \textbf{53.71} & \textbf{57.95}
\end{tabular}}%
\end{center}
\vspace{-0.5em}
\caption{\label{tab:lm} Perplexity of MT output on newstest2018 based on LMs trained on natural text vs. translated text, binned by adequacy. Sup and Unsup are comparable supervised and unsupervised MT systems, respectively. $\downarrow$ from the Natural Text LM and $\uparrow$ from the Translated Text LM indicate more natural-sounding output.} 
\vspace{-1.25em}
\end{table*} 

\paragraph{Stronger Supervised MT} Though analyzing systems of similar quality is important for head-to-head comparison, we evaluate a stronger supervised system for context.\footnote{Trained on 4.5 million lines of WMT14 bitext.} We do not have human evaluation scores, but automatic results give insight: see Table \ref{tab:stronger_sup}. The model has overall BLEU = 40.9 and a similarly large discrepancy on orig-en vs. orig-de as did the Sup system used throughout this work: 44.6 for orig-en and 34.9 for orig-de. 
As for structural similarity, this stronger system has lower overall posTER vs. the human reference---0.238 vs. 0.280/0.287 from Sup/Unsup---indicating even more structural similarity with the reference. For naturalness, the stronger system has lower perplexity from the nLM. As a higher-quality system, this is expected. At the same time, it scores much lower than Sup and Unsup by the tLM, where higher indicates more natural-sounding output: 29.23 vs. 41.06/58.17 for Sup/Unsup.

\begin{table}[h]
\vspace{-0.25em}
\small
\begin{center}
\resizebox{\columnwidth}{!}{%
\begin{tabular}{rr|rr|rr}
\toprule
\multicolumn{2}{c}{\bf Quality} & \multicolumn{2}{c}{\bf Structural Sim.} & \multicolumn{2}{c}{\bf Naturalness} \\
 BLEU & nt18p & v. Src & v. Ref & nLM & tLM \\
\midrule
 40.9 & 12.1 & 0.401 & 0.238 & 54.35 & 29.23 \\
\bottomrule
\end{tabular}}%
\end{center}
\vspace{-0.5em}
\caption{\label{tab:stronger_sup} Strong supervised model trained on WMT14. Structural sim. is posTER: v. Src is comparable to Table \ref{tab:ter_cross}, v. Ref to \textit{Overall} in Table \ref{tab:ter_qual}. $\downarrow$ = more monotonic. nLM/tLM are Natural/Translated Text LMs of Table \ref{tab:lm}.}
\vspace{-1.5em}
\end{table}

\paragraph{Ablation: Architecture vs. Data}

\begin{table*}[hbt]
\small
\begin{center}
\resizebox{\textwidth}{!}{%
\setlength{\tabcolsep}{8pt}
\begin{tabular}{c||c|c||c|c|c}
 & \multicolumn{2}{c||}{\bf LM Perplexity (PPL)} & \multicolumn{3}{c}{\bf BLEU} \\
\hline
& \bf Natural Text LM & \bf Translated Text LM & \bf Overall & \bf orig-en & \bf orig-de \\ 
\hline
\bf Supervised (Sup)                & 72.69         & 41.06  & 29.2 & 34.0 & 21.1  \\
\bf Sup En-Trns/De-Orig  & 69.75 &         50.65  & 35.4 & 35.5 & 34.1  \\
\bf Unsup                & \bf 67.01 & \bf 58.17  & 30.1 & 30.9 & 27.1  \\
\bf Unsup-Trns           & 69.88 &         48.90  & 33.4 & 35.4 & 28.4
\end{tabular}}%
\end{center}
\vspace{-0.5em}
\caption{\label{tab:arch_v_data_sys} Comparison of 4 English$\to$German MT systems: ppl from LMs trained on natural or translated text, BLEU on newstest2018. $\downarrow$ ppl = model prefers the output. Sup En-Trns/De-Orig is supervised, trained on translated English and German-original News Crawl. Unsup is trained on natural English and German News Crawl. Unsup-Trns uses translated News Crawl only. Unsup performs best == more like natural text and less like translated text.}
\vspace{-0.5em}
\end{table*}

One reason Unsup might produce more natural-sounding output could be simply that it develops language-modeling capabilities from natural German alone, whereas Sup must see some translated data (being trained on bitext of human translations). Next, we ask whether the improved naturalness and structural diversity is due to the unsupervised NMT architecture, or simply the natural training data. 

We build a supervised en-de MT system with 329,000 paired lines of translated English source and natural German, where the source is back-translated German News Crawl from a supervised system. In other words, we train on backtranslated data only on the source side and natural German as the target. The model thus develops its language-modeling capabilities on natural sentences alone. If more natural output is simply a response to training on natural data, then this supervised system should perform as well in naturalness as Unsup, or better.

We train another unsupervised system on translated text only. Source-side training data is synthetic English from translating German News Crawl with a supervised system. Target-side is synthetic German which was machine-translated from English News Crawl. If naturalness solely results from data, this system should perform worst, being trained \textit{only} on translated (unnatural) text.  

Table \ref{tab:arch_v_data_sys} shows the results. The original unsupervised system (Unsup) performs best according to both LMs, having output that is more natural and less like translated text. When given only natural German to build a language model, the supervised system (Sup En-Trns/De-Orig) \textit{still} produces more unnatural output than Unsup. Even when the unsupervised system uses translated data only (Unsup-Trns), its output is \textit{still} more natural than the original supervised system (Sup) according to both LMs. This is a surprising result, and is interesting for future study. Together, these findings suggest that both German-original data \textit{and} the unsupervised architecture encourage output to sound more natural. 

\section{Application: Leveraging Unsupervised Back-translation}
\label{sec:joined_mt}

Our results indicate that high-adequacy unsupervised MT output is more natural and more structurally diverse in comparison to human translation, than is supervised MT output. We are thus motivated to use these advantages to improve translation.
We explore how to incorporate unsupervised MT into a supervised system via back-translation. We train for $\sim$500,000 updates for each experiment, and select models based on validation performance on newstest2018. We test on newstest2019(p).

\subsection{Baselines}
The first row of Table \ref{tab:sup_plus_unsup} is the supervised baseline trained on the WMT18 bitext. The second row is Unsup, used throughout this work.

We back-translate 24 million randomly-selected sentences of German News Crawl twice: once using a supervised German-English system trained on WMT18 bitext with a transformer-big architecture, and once using Unsup. Both use greedy decoding for efficiency. We augment the WMT18 bitext with either the supervised or unsupervised BT. 

\begin{figure*}[h]
  \centering
  \includegraphics[height=0.17\textheight, width=0.80\linewidth]{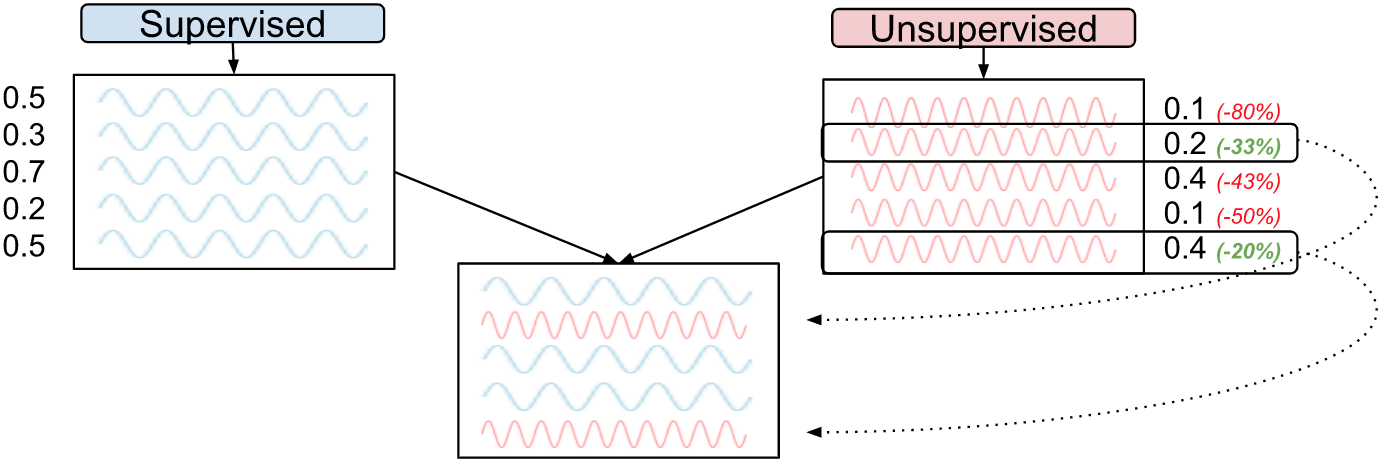}
  \caption{Back-translation selection method. Both systems translate the same source sentences. If an unsupervised output sentence is more than T\% as likely as the supervised one, select the unsupervised. Here, T=65\%.}
  \label{fig:bt-selection}
\vspace{-1.5em}
\end{figure*}

Seen in Table \ref{tab:sup_plus_unsup}, adding supervised BT (+SupBT) performs as expected; minorly declining on the source-original test set (orig-en), improving on the target-original set (orig-de), and improving on the paraphrase set (nt19p). Conversely, adding unsupervised BT (+UnsupBT) severely lowers BLEU on source-original and paraphrase test sets. Randomly-partitioning the BT sentences such that 50\% are supervised BT and 50\% are unsupervised also lowers performance on orig-en (+50-50BT).

\begin{table*}[htb]
\small
\begin{center}
\resizebox{\textwidth}{!}{%
\setlength{\tabcolsep}{8pt} %default is 6pt
\begin{tabular}{@{}l||c|c|c|c||c|c|c@{}}
& \multicolumn{4}{c||}{\bf{newstest2018}} & \multicolumn{3}{c}{\bf{newstest2019}} \\
\hline
& \bf Overall & \bf orig-en & \bf orig-de & \bf nt18p &\bf orig-en & \bf orig-de & \bf nt19p \\
\hline
Supervised Baseline (5.2M) &  41.8 &	46.1 &	34.3 & 12.6 & 38.8 & 30.4 & 11.7 \\
Unsup \textit{(used throughout this work)} & 30.1 & 30.9 &	27.1 & 9.6 & 24.6 & 28.5 & 8.8 \\
\hline
\it Supervised Baseline & & & & & & \\
\hspace{3mm}+ SupBT & 43.4 & 43.7 &	41.8 & 12.5 & 37.0 & \bf 39.9 & 12.0 \\
\hspace{3mm}+ UnsupBT & 33.3 &	33.8 &	31.1 & 9.9 & 27.2 & 30.8 & 9.5 \\
\hspace{3mm}+ 50-50BT & 38.0 &	36.4 &	39.0 & 12.9 & 29.4 & 38.3 & 10.0 \\
\hline
\hspace{3mm}+ SupBT\_Tag & \bf 44.8 &	47.0 &	40.7 & 13.0 & \bf 40.3 & 38.2 & 12.4 \\
\hspace{3mm}+ UnsupBT\_Tag & 43.3 &	46.9 &	36.9 & 12.9 & 39.1 & 35.0 & 12.2 \\
\hspace{3mm}+ 50-50BT\_Tag & 44.4	 & \bf 47.1	& 39.6 & 12.9 & 39.4 & 38.0 & 12.2 \\ 
\hspace{3mm}+ 50-50BT\_TagDiff & 44.4 & 46.8 & 40.1 & 13.0	& 39.9	& 37.9	& 12.4 \\ 
\hline
\hspace{3mm}+ SmallMix\_Tag & \bf 44.8	& 46.8	& 40.8 & \bf 13.2 & 39.8 & 38.8 & 12.5 \\
\hspace{3mm}+ MediumMix\_Tag  & 44.7 & 46.8 &	40.8 & 13.0 & 40.1 & 38.2 & \bf 12.6
\end{tabular}}%
\end{center}
\vspace{-0.5em}
\caption{\label{tab:sup_plus_unsup} SacreBLEU of supervised baseline plus 24M supervised or unsupervised BTs. +MediumMix\_Tag and +SmallMix\_Tag use the BT selection method of \S\ref{sec:bt-selection}. +MediumMix\_Tag has 9.4M unsupervised BT and 14.6M supervised BT. +SmallMix\_Tag has 3.1M and 20.9M, respectively. nt18p and nt19p are paraphrase references from~\citet{freitag-etal-2020-bleu}, where small BLEU score changes can indicate tangible quality difference.} 
\vspace{-1.25em}
\end{table*}

\subsection{Tagged BT}
Following~\citet{caswell-etal-2019-tagged}, we tag BT on the source-side. Tagging aids supervised BT (+SupBT\_Tag) and greatly improves unsupervised BT (+UnsupBT\_Tag), which outperforms the baseline and is nearly on-par with +SupBT\_Tag. Combining supervised and unsupervised BT using the same tag for both (+50-50BT\_Tag) shows no benefit over +SupBT\_Tag. +50-50BT\_TagDiff uses different tags for supervised vs. unsupervised BT. 

\subsection{Probability-Based BT Selection}
\label{sec:bt-selection}
We design a BT selection method based on translation probability to exclude unsupervised BT of low quality. We assume that supervised BT is ``good enough.'' Given translations of the same source sentence (one supervised, one unsupervised) we assert that an unsupervised translation is ``good enough'' if its translation probability is similar or better than that of the supervised translation. If much lower, the unsupervised output may be low-quality. 
\begin{itemize}
[itemsep=3pt,topsep=3pt]
\item Score each supervised and unsupervised BT with a supervised de-en system.
\item Normalize the translation probabilities to control for translation difficulty and output length.
\item Compare probability of the supervised and unsupervised BT of the same source sentence: 
\begin{equation*}
    \setlength{\abovedisplayskip}{3pt}
    \setlength{\belowdisplayskip}{1.5pt}
    \Delta P = \frac{P \text{norm}(\text{unsup})}{P \text{norm}(\text{sup})}
\end{equation*}
\item Sort translation pairs by $\Delta$P.
\item Select the unsupervised BT for pairs scoring highest $\Delta$P and the supervised BT for the rest.
\end{itemize}
This filters out unsupervised outputs less than a hyperparameter T\% as likely as the corresponding supervised sentence and swaps them with the corresponding supervised sentence. Importantly, the same 24M source sentences are used in all experiments. The procedure is shown in Figure \ref{fig:bt-selection}.

Full results varying T are in the Appendix for brevity, but we show two example systems in Table \ref{tab:sup_plus_unsup}. The model we call ``+MediumMix\_Tag'' uses the top $\sim$40\% of ranked unsupervised BT with the rest supervised (9.4M unsupervised, 14.6M supervised).  ``+SmallMix\_Tag'' uses the top $\sim$13\% of unsupervised BT (3.1M unsupervised, 20.9M supervised).\footnote{The numbers are not round because data was selected using round numbers for the hyperparameter T.} We use the same tag for all BTs. Improvements are modest, but our goal was to demonstrate how one might use unsupervised MT output rather than build a state-of-the-art system.

+SmallMix\_Tag performs better than the previous best on newstest2018p and +MediumMix\_Tag performs highest overall on nt19p. We recall that small differences on paraphrase test sets can signal tangible quality differences \cite{freitag-etal-2020-bleu}. Trusting BLEU on nt19p, we use \emph{+MediumMix\_Tag} as our model for human evaluation.

One might inquire whether improved performance is due to the simple addition of noise in light of \citet{edunov2018understanding}, who conclude that noising BT improves MT quality. Subsequent work, however, found that benefit is not from the noise itself but rather that noise helps the system distinguish between  parallel and synthetic data \cite{caswell-etal-2019-tagged, marie-etal-2020-tagged}. \citet{yang2019effectively} also propose tagging to distinguish synthetic data. With tagging instead of noising, \citet{caswell-etal-2019-tagged} outperform \citet{edunov2018understanding} in 4 of 6 test sets for En-De, furthermore find that noising on top of tagging does not help. They conclude that ``tagging and noising are not orthogonal signals but rather different means to the same end.'' In light of this, our improved results are likely not due to increased noise but rather to systematic differences between supervised and unsupervised BT. 

\subsection{Human Evaluation}
We run human evaluation with professional translators for \emph{+MediumMix\_Tag}, comparing its output translation of the newstest2019 test set with two baseline models. Table \ref{tab:final_human_eval} shows that humans prefer the combined system over the baseline outputs.\footnote{Scores are low because we use only WMT18 bitext + BT, and translators score more harshly than crowd workers.} 
Table \ref{tab:fluency} shows the percentage of sentences judged as ``worse than,'' ``about the same as,'' or ``better than'' the corresponding +SupBT\_Tag output, based on fluency. Raters again prefer the combined system. The improvements are modest, but encouragingly indicate that unsupervised MT may have something to contribute to machine translation, even in high-resource settings.

\begin{table}[]
\vspace{-0.25em}
\small
\begin{center}
\resizebox{\columnwidth}{!}{%
\setlength\tabcolsep{20pt} % default value: 6pt
\begin{tabular}{l|c}
& \bf Adequacy \\ 
\hline
+ UnsupBT\_Tag & 54.82 \\
+ SupBT\_Tag & 56.13 \\
+ MediumMix\_Tag & \bf 58.62
\end{tabular}}%
\end{center}
\vspace{-0.5em}
\caption{\label{tab:final_human_eval} Human-eval direct assessment (adequacy) of supervised MT with supplemental back-translation.}
\vspace{-0.75em}
\end{table}

\begin{table}[]
\small
\begin{center}
\resizebox{\columnwidth}{!}{%
\setlength\tabcolsep{22pt} % default value: 6pt
\begin{tabular}{c|c|c}
\hline
Better & Same & Worse  \\
\hline
51.1\% &  3.7\% & 45.2\%  
\end{tabular}}%
\end{center}
\vspace{-0.5em}
\caption{\label{tab:fluency} Human side-by-side fluency eval. Shown: \% of +MediumMix\_Tag sentences judged ``worse than,'' ``about the same,'' or ``better than'' +SupBT\_Tag output.}
\vspace{-1.5em}
\end{table}

\section{Conclusion}
\label{sec:conclusion}
Recent unsupervised MT systems can reach reasonable translation quality under clean and controlled data conditions,
and could bring alternative translations to language pairs with ample parallel data. We perform the first systematic comparison of supervised and unsupervised MT output from systems of similar quality. We find that systematic differences do exist, and that high-quality unsupervised MT output appears \textbf{more natural} and \textbf{more structurally diverse} when compared to human translation, than does supervised MT output. Our findings indicate that there may be useful differences between supervised and unsupervised MT systems that could contribute to a system better than either achieves alone. As a first step, we demonstrate an unsupervised back-translation augmented model that takes advantage of the differences between the translation methodologies to outperform a traditional supervised system on human-evaluated measures of adequacy and fluency. 

\bibliography{anthology,custom}
\bibliographystyle{acl_natbib}

\appendix

\begin{table*}[]
\small
\setlength\tabcolsep{2pt} % default value: 6pt
\begin{center}
\begin{tabular}{l||c|c|c|c||c|c|c}
\hline
& \multicolumn{4}{c||}{\bf{newstest2018}} & \multicolumn{3}{c}{\bf{newstest2019}} \\
\hline
& \bf Joint & \bf Orig-En & \bf Orig-De & \bf nt18p &\bf Orig-En & \bf Orig-De & \bf nt19p \\
\hline
Supervised Baseline (5.2M) &  41.8 &	46.1 &	34.3 & 12.6 & 38.8 & 30.4 & 11.7 \\
Unsup \textit{(same used throughout this work)} & 30.1 & 30.9 &	27.1 & 9.6 & 24.6 & 28.5 & 8.8 \\
\hline
\it Supervised Baseline & & & & & & \\
\hspace{3mm}+ SupBT & 43.4 & 43.7 &	41.8 & 12.5 & 37.0 & \bf 39.9 & 12.0 \\
\hspace{3mm}+ UnsupBT & 33.3 &	33.8 &	31.1 & 9.9 & 27.2 & 30.8 & 9.5 \\
\hspace{3mm}+ 50-50BT & 38.0 &	36.4 &	39.0 & 12.9 & 29.4 & 38.3 & 10.0 \\
\hline
\hspace{3mm}+ SupBT\_Tag & \bf 44.8 &	47.0 &	40.7 & 13.0 & \bf 40.3 & 38.2 & 12.4 \\
\hspace{3mm}+ UnsupBT\_Tag & 43.3 &	46.9 &	36.9 & 12.9 & 39.1 & 35.0 & 12.2 \\
\hspace{3mm}+ 50-50BT\_Tag & 44.4	 & \bf 47.1	& 39.6 & 12.9 & 39.4 & 38.0 & 12.2 \\ 
\hspace{3mm}+ 50-50BT\_TagDiff & 44.4 & 46.8 & 40.1 & 13.0	& 39.9	& 37.9	& 12.4 \\
\hline
\hspace{3mm}+ 21.7M Tagged Unsup \& 2.3M Sup BT & 44.0	& 46.6 & 39.3 & 13.0 & 39.6 & 36.9 & 12.3 \\
\hspace{3mm}+ 17.4M Tagged Unsup \& 6.6M Sup BT & 44.0	& 46.2 & 40.0 & 13.0 & 40.0 & 37.7 & 12.3 \\
\hspace{3mm}+ 9.4M Tagged Unsup \& 14.6M Sup BT (``+MediumMix\_Tag'') & 44.7 & 46.8 &	40.8 & 13.0 & 40.1 & 38.2 & \bf 12.6 \\
\hspace{3mm}+ 3.1M Tagged Unsup \& 20.9M Sup BT (``+SmallMix\_Tag'') & \bf 44.8	& 46.8	& 40.8 & \bf 13.2 & 39.8 & 38.8 & 12.5 \\
\hspace{3mm}+ 1.5M Tagged Unsup \& 22.5M Sup BT & \bf 44.8	& \bf 47.1	& 40.7 & \bf 13.2 & 40.0 & 38.4 & 12.5 \\
\hspace{3mm}+ 680K Tagged Unsup \& 23.3M Sup BT & 44.4 & 46.4 &	40.7 & 12.9 & 40.0 & 38.1 & 12.4 \\
\hline
\end{tabular}
\end{center}
\caption{\label{tab:sup_plus_unsup_addl} SacreBLEU of supervised baseline plus 24M supervised or unsupervised BTs. Systems using both use the BT selection method of \S\ref{sec:bt-selection} with increasing values for hyperparameter $T$. nt18p and nt19p are paraphrase references from~\citet{freitag-etal-2020-bleu}, where small BLEU score changes can indicate tangible quality difference.} 
\end{table*}

\end{document}